\documentclass[journal]{IEEEtran}
\usepackage{times}
\usepackage{epsfig}
\usepackage{graphicx}
\usepackage{amsmath}
\usepackage{amssymb}

\begin{document}

\title{Learning landmark guided embeddings for animal re-identification}

\author{Olga~Moskvyak, Frederic~Maire, Feras~Dayoub and~Mahsa~Baktashmotlagh
\thanks{O. Moskvyak, F. Maire and F. Dayoub are with the School of Electrical Engineering 
and Computer Science, Queensland University of Technology, Brisbane, QLD 4000, Australia.}
\thanks{M.~Baktashmotlagh is with the School of Information Technology and Electrical Engineering, The University of Queensland, St Lucia, QLD 4072, Australia.}%
\thanks{Corresponding author O. Moskvyak: olga.moskvyak@hdr.qut.edu.au}}

\maketitle

\begin{abstract}
   Re-identification of individual animals in images can be ambiguous due to subtle variations in body markings between different individuals and no constraints on the poses of animals in the wild.
   Person re-identification is a similar task and it has been approached with a deep convolutional neural network (CNN) that learns discriminative embeddings for images of people.
   However, learning discriminative features for an individual animal is more challenging than for a person's appearance due to the relatively small size of ecological datasets compared to labelled datasets of person's identities.
   We propose to improve embedding learning by exploiting body landmarks information explicitly.
   Body landmarks are provided to the input of a CNN as confidence heatmaps that can be obtained from a separate body landmark predictor.
   The model is encouraged to use heatmaps by learning an auxiliary task of reconstructing input heatmaps.
   Body landmarks guide a feature extraction network to learn the representation of a distinctive pattern and its position on the body.
   We evaluate the proposed method on a large synthetic dataset and a small real dataset.
   Our method outperforms the same model without body landmarks input by 26\% and 18\% on the synthetic and the real datasets respectively.
   The method is robust to noise in input coordinates and can tolerate an error in coordinates up to 10\% of the image size.
   
\end{abstract}

\section{Introduction}

Animal re-identification in images is an instance level recognition and retrieval problem which aims to distinguish between individual animals and find matching examples in an image database.
Individual animals can be told apart by subtle variations in natural markings on their body such as belly patterns on manta rays, stripes on tigers and zebras.

Automatic re-identification of animals in photos is of high importance for wildlife monitoring and conservation because it is less time consuming than manual visual inspection and more efficient than collecting biology samples or attaching and tracking microchips \cite{into-photo-id-rays-sharks}.

\begin{figure}
\begin{center}
 \includegraphics[width=\linewidth]{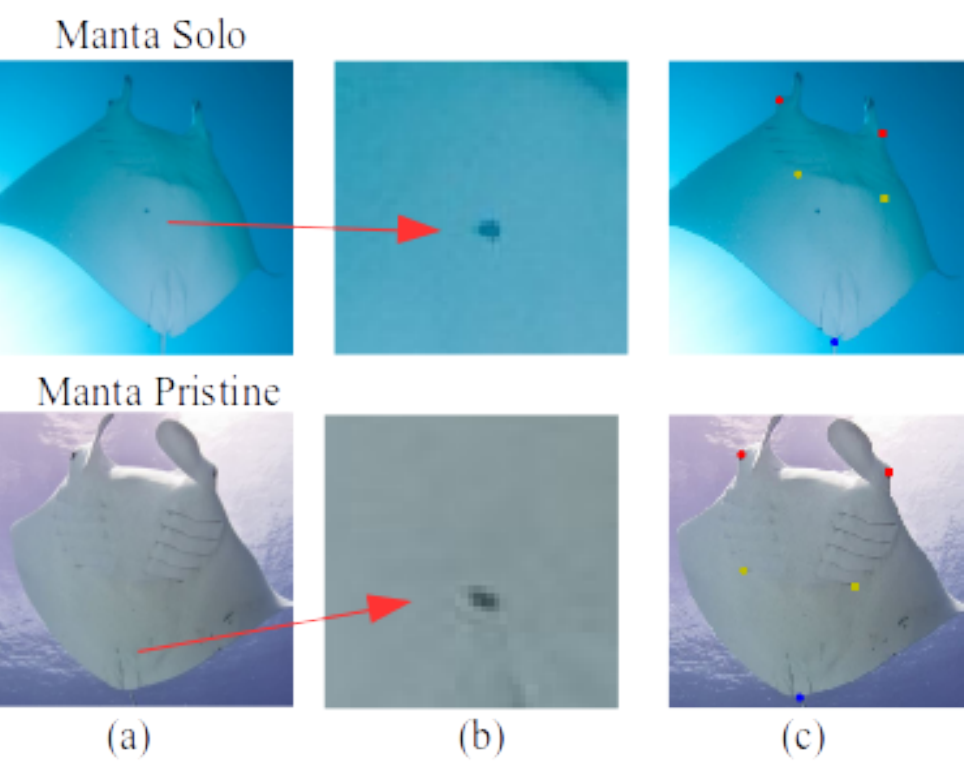}
\end{center}
   \caption{(a) The spot pattern on these two different manta rays is the same (consists of one black dot). 
   (b) Localised images of the spot pattern are ambiguous. 
   (c) To distinguish between individuals like these we propose to exploit body landmark coordinates (e.g., eyes, gills, a tail) in the re-identification system. Photo credit: David Biddulph, John Gransbury.}
\label{fig:intro_image}
\end{figure}

\begin{figure*}
\begin{center}
 \includegraphics[width=0.85\linewidth]{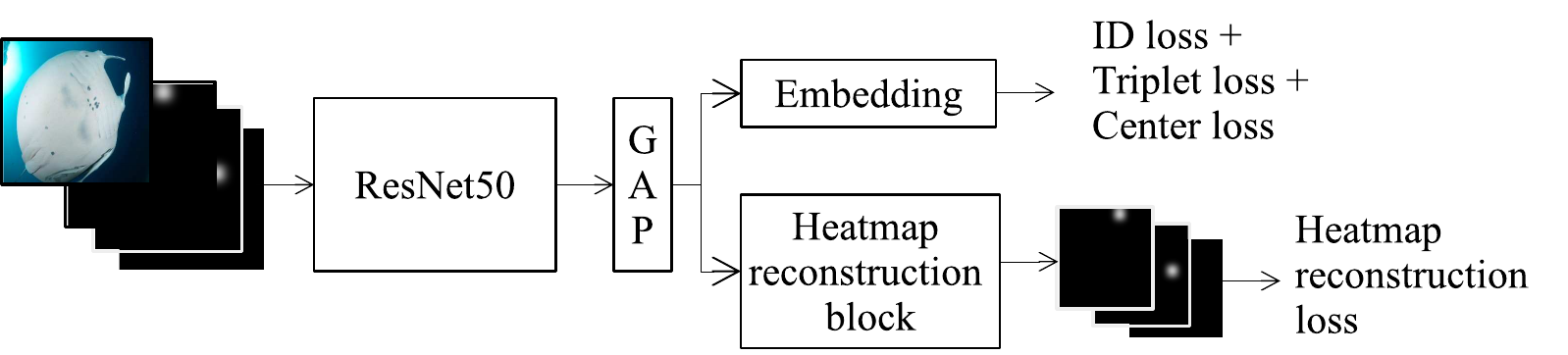}
\end{center}
   \caption{Architecture of our \textit{Landmark-Id} model. The novel features are additional heatmap input and heatmap reconstruction block. GAP is global average pooling of feature maps.}
\label{fig:model_architecture}
\end{figure*}

The task is similar to person recognition that has been approached with deep convolutional neural networks \cite{person-review}.
A network learns embeddings for images of people's appearances in such a way that the distance between embeddings of the same person is smaller than the distance between embeddings  of  different people.
However, visual animal re-identification is mainly based on body markings which are more ambiguous than a person’s appearance because of the similarities among different individuals. 
For example, different manta rays can have a very similar spot pattern but located at different positions on the belly.
Figure~\ref{fig:intro_image} shows two manta rays with only one black dot on the belly and the only difference is the location of the spot with respect to the landmarks (e.g., eyes, a base of the tail and gills).

There are limitations in transferring face and person re-identification methods to images of animals:
\begin{itemize}
    \item faces are usually normalised to an upright frontal pose thanks to robust methods to detect facial landmarks and body postures are aligned vertically;
    \item warping to a canonical position or alignment is not always possible for animals due to the sensitivity of these methods to errors in coordinates of body landmarks;
    \item wildlife datasets have limited data compared to large public datasets for face and person re-identification so there is less chance to learn the relation between the body landmarks and unique markings from the data itself.
\end{itemize}

Previous work on the manta ray re-identification system \cite{manta-reid} uses cropped images of spot patterns to focus the model's attention on the pattern itself and avoid distraction from the background.
However, the cropped patch of a sport pattern loses information about its relative position on the body
so it is not likely to correctly identify individuals with similar patterns that differ only in a position like in Figure~\ref{fig:intro_image}.

We build on a strong model for person re-identification \cite{person-reid-baseline} and propose to improve embedding learning for animal re-identification
by adding locations of body landmarks to the model input.
The new model explicitly receives information about the position of distinctive features with respect to the body.
The motivation of using body landmarks is the scarcity of annotated datasets with animal identities compared to large labelled datasets for person re-identification.
Identification based only on the pattern itself without the knowledge about the position of a specific mark  can be error prone.
We favor heatmaps over exact coordinates to encode the estimated body landmark location because heatmaps can represent uncertainty.

The key contributions of this paper are:
\begin{itemize}
    \item a novel method to exploit body landmark locations to improve the performance of re-identification system;
    \item a novel heatmap augmentation method to train the model to handle missing or not visible landmarks;
    \item robustness to uncertainty in body landmark coordinates up to 10\% of the image size.
\end{itemize}

\section{Related work}

\begin{figure*}[t!]
\begin{center}
 \includegraphics[width=0.9\linewidth]{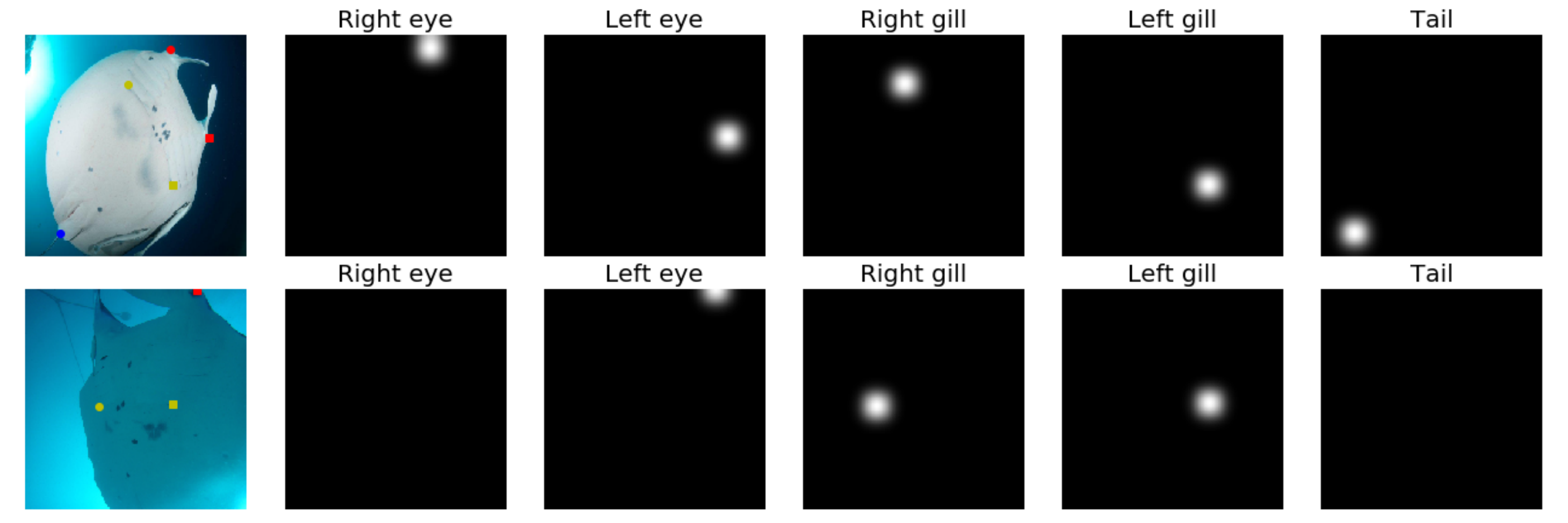}
\end{center}
   \caption{Example of heatmaps for two images. One grayscale channel per a landmark. If the landmark is not visible, the heatmap is black. These heatmaps have the bright blob with the radius of 5\% of the image size.}
\label{fig:manta_heatmaps}
\end{figure*}

There are multiple approaches to re-identification and some of them use only pixel intensities and some leverage additional information such as the semantic structure of the object (e.g., body parts of a person).
We discuss re-identification methods that include some degree of pose or body landmark information.
Pose information can be exploited to align the object of interest to a standard pose or crop patches from the image to obtain local features.

The body and face alignment based on keypoints is used to eliminate pose variance and improve recognition performance. 
Zheng et al.~\cite{person-posebox} introduce the PoseBox structure to align pedestrians to a standard pose. 
The alignment is used extensively for face recognition \cite{parkhi2015deep, schroff2015facenet, sun2014deep}.
Normalizing the head orientation of right whales improves the re-identification performance \cite{whale-re-id}.
However, accurate body landmark information is hard to obtain and alignment methods are sensitive to precise coordinates \cite{multy-view-geometry}.
It is not feasible to transfer alignment methods directly from person to animal re-identification.

Landmark coordinates are used to extract local features from various patches cropped from the image.
Recognition and fine-grained classification methods use these local features to complement a global representation.
Guo and Farrell \cite{birds-pose} construct object representation as the concatenation of hierarchical pose-aligned regions features extracted from patches around pairs of body landmarks.
Tiger re-identification has been improved by concatenating global features extracted from the whole image with local features from limb's patches \cite{tiger-part-pose-guided}. 
Su et.~al \cite{person-posedriven} explicitly leverage human body part cues to detect and normalize body parts to extract local features and combine them in a pose driven feature weighting subnetwork.

Pose information has also been used to enhance re-identification by generating new data samples in a pose-transferable person re-id framework \cite{person-transferrable-reid}.
However, training image generation models requires a large amount of data so it cannot be transferred directly to smaller datasets of animal identities.
    
Several works include pose information to guide feature extraction.
Sarfraz et.~al \cite{pose-guided-person} improve person re-identification by incorporating both fine and coarse pose information into learning discriminative embeddings.
Fine pose information is confidence maps from off-the-shelf body landmark predictor.
Coarse pose information is the quantization (‘front’, ‘back’, ‘side’) of a person’s orientation to the camera.
Liu et.~al \cite{tiger-pose-guided} simplifies the tiger body pose into two categories according to the heading direction of the tigers to reduce pose variations.
However, these approaches are not transferable to other objects due to coarse pose labels are specific to the task.
In this paper, we introduce a generic method of exploiting body landmark information to improve learning of discriminative embeddings.

\section{Learning landmark guided embeddings}

A pose of an animal's body with respect to the camera greatly affects the appearance of natural markings and the visibility of body landmarks in the photo.
It is hard to obtain accurate locations of body landmarks because images are taken in the wild environment with an unknown pose of the animal in front of the camera, complex natural backgrounds and changing lighting conditions.
Information about the pose has the potential to improve re-identification performance.

\subsection{Baseline re-identification model}
As a baseline re-identification model, we use the second best model developed for person re-identification \cite{person-reid-baseline} that is generic enough to be transferred from people to animals.
The state-of-the-art for person re-identification requires spatial-temporal information \cite{person-reid-sota} and cannot be transferred to our task.

The backbone of the baseline model is ResNet50 \cite{resnet} that is initialized with pre-trained parameters on ImageNet.
The model outputs ReID features $f$ and ID prediction logits $p$.
ReID features $f$ are used to calculate a triplet loss \cite{triplet-loss} and a center loss \cite{center-loss}. 
Triplet loss pulls embeddings of images from the same individual closer together while pushing embeddings of images of different individuals above a specified margin.
Center loss penalizes the distance between embeddings and their corresponding class centers where each individual is a class.
ID prediction logits $p$ are used to calculated a smoothed cross entropy loss \cite{smooth-ce}  over training classes  to facilitate learning of discriminative features and are discarded at inference. 
The training process and all hyperparameters are inherited from the original work \cite{person-reid-baseline}. 
The baseline model is optimized with a weighted combination of three losses: the smoothed cross-entropy $L_{\textnormal{ID}}$ over training classes, the triplet loss $L_{\textnormal{Triplet}}$ and the center loss $L_{\textnormal{Center}}$:

\begin{equation}
    L_{\textnormal{ReId}} = L_{\textnormal{ID}} + L_{\textnormal{Triplet}} + \beta L_{\textnormal{Center}}
\end{equation}
where $\beta = 0.0005$ as in \cite{person-reid-baseline}.

\subsection{Landmark aware re-identification model}

We add the body landmark information to the model input by concatenating $k$ extra channels with three RGB image channels ($k$ is a number of landmarks).
Each channel is a grayscale heatmap representing the likelihood of a landmark location.
Figure~\ref{fig:manta_heatmaps} shows two images of manta rays and corresponding heatmaps.
These additional channels guide feature extraction to learn embeddings that are aware of the location of distinctive features with respect to body parts.

Information about landmarks can be obtained from another model that predicts landmarks based on the image (this task is out of the scope of the current work).
Landmarks can also be annotated manually.

The model is encouraged to use landmark information by learning the auxiliary reconstruction task of input heatmaps from embeddings, see Figure~\ref{fig:model_architecture}. 
We also experimented with the heatmap reconstruction block branched off after the third dimensionality reduction step and the results were similar to the reconstruction from the final features.
We call this model a \textit{Landmark-Id} model.

The \textit{Landmark-Id} model is optimised with the following loss:

\begin{equation}
    L = L_{\textnormal{ReId}} + \alpha L_{\textnormal{HR}} 
\end{equation}
where heatmap reconstruction loss $L_{\textnormal{HR}}$ is a binary cross-entropy.
We experimented with $\alpha$ equal to 0.1, 1 and 10 and observed no difference in accuracy so we set $\alpha = 1$. 

The \textit{Landmark-Id} model is trained in two stages. 
At the first stage only randomly initialised weights in the first layer and in the final classification layer are trained while all other weights remain fixed.
Once these layers are adapted to the rest of the network, the whole network is fine-tuned.
The parameters in the first layer of ResNet50 are initialized randomly because the number of input channels differs from the number of channels in ImageNet due to the additional heatmap input.
The rest of the parameters in ResNet50 model are initialized with ImageNet pretrained weights.
Heatmap reconstruction block is not trained at this stage.

At the second stage, the heatmap reconstruction branch is added with randomly initialised weights. Only the heatmap reconstruction block is trained for the first ten epochs to tune random weights. Then the whole model is fine-tuned with a ten times smaller learning rate than in the first stage.

\subsection{Heatmap augmentation}

\begin{figure}
\begin{center}
 \includegraphics[width=\linewidth]{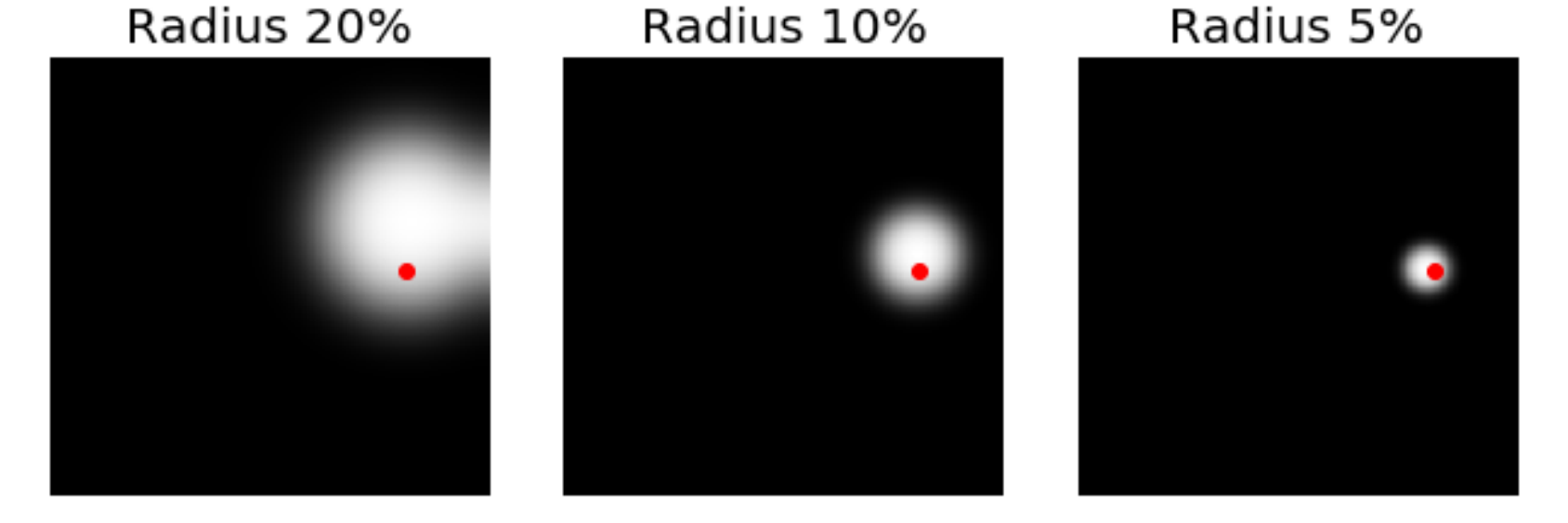}
\end{center}
   \caption{Noisy landmark augmentation (NLA) on heatmaps with different levels of uncertainty about landmark locations. The true coordinate is located inside the bright blob but not necessary in the middle.}
\label{fig:heatmaps_size_noise}
\end{figure}

We introduce two augmentation techniques for heatmaps to improve the generalization ability of the \textit{Landmark-Id}: noisy landmark augmentation (NLA) and missing landmark augmentation (MLA).
Locations of body landmarks cannot always be annotated correctly especially when these are obtained from an automated landmark detection method.
Due to large variations in animal poses some landmarks may not be visible in the image. 
NLA and MLA address these two problems.

NLA randomly shifts the blob in each heatmap (by default the center of the blob is the landmark) by a number of pixels less or equal than the radius of the blob, see Figure~\ref{fig:heatmaps_size_noise}.
This way the landmark location is still contained within a blob but not always in the middle.

MLA has two parameters: a minimum number $M$ of visible landmarks (specific to the dataset) and a probability $p_{\textnormal{mla}}$. If there are more than $M$ landmarks visible in the image, than some of them may be set to missing with probability $p_{\textnormal{mla}}$. In practice, a missing landmark means that the corresponding heatmap is set to all zeros.
The motivation for this augmentation is imbalanced data when there are not enough examples for the model to learn to reconstruct black heatmaps for not visible landmarks.
We list the hyperparameters used for MLA in the Experiments section.

\section{Experiments}

\subsection{Datasets}

\subsubsection{Synthetic dataset}

\begin{figure}
\begin{center}
 \includegraphics[width=\linewidth]{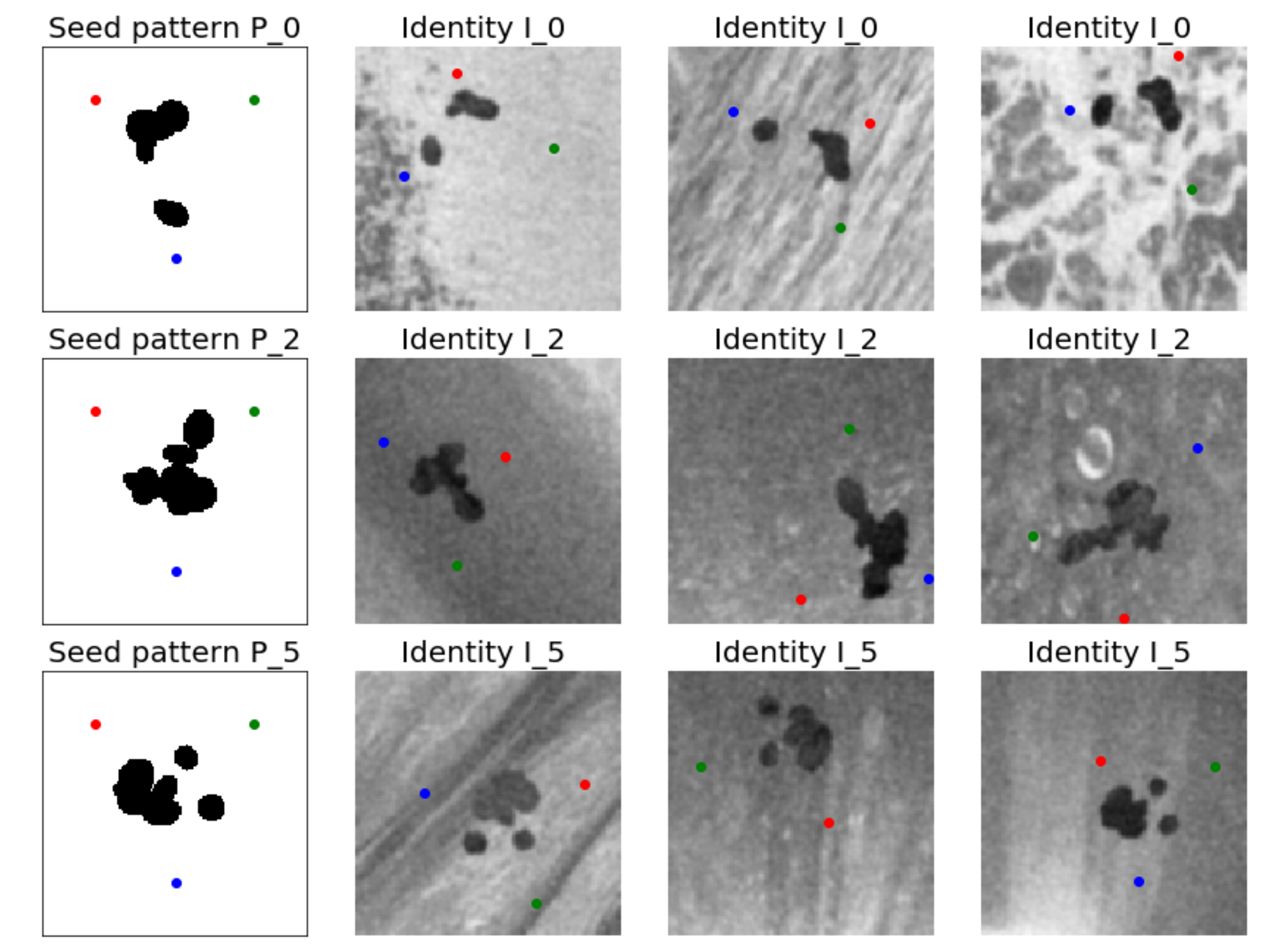}
\end{center}
   \caption{Example of three identities from the synthetic dataset. Each row shows a seed pattern and three generated examples for one identity. Coloured points represent locations of three landmarks and are plotted over images for illustration only and do not appear on images in the generated dataset.}
\label{fig:synthetic_patterns}
\end{figure}

We verify ideas on a synthetic dataset first as it gives the ability to control the number and variety of examples.
The design of synthetic images is inspired by manta rays belly patterns but does not aim to replicate it.
Consider a collection of seed patterns \( \mathcal{P} = \{P_1, \ldots, P_n\}\)  where each pattern \( P_{i} \) is a unique pattern of black filled ellipses on a white background inside at triangle area in the center as illustrated in Figure~\ref{fig:synthetic_patterns} (first column). 
The corners of the triangle play the role of body landmarks.
The image itself does not have any information about landmark locations.
Landmark coordinates are recorded in a separate array.

Seed patterns represent a canonical view from a camera placed directly in front of it. 
When the camera moves, the projection of the pattern on the camera plane will be related to the  canonical view by a homography.
We call an \textit{identity} $I_i$ a unique pattern of ellipses where examples belonging to this identity are generated by applying random projective transformations 
to the canonical pattern $P_i$ and adding a random background (see examples in Figure~\ref{fig:synthetic_patterns}).
To randomise textures of a background and a pattern we use patches from images showing underwater scenes without any salient objects \cite{unsplash}.
The pixel intensities in a background image are rescaled to be lighter than pixel intensities of a pattern texture to avoid merging of the pattern with the background.
Finally, images are converted to grayscale and Gaussian noise is added.
Landmark coordinates are warped the same way as a pattern and are recorded in a corresponding array.


The dataset consists of 3 subsets: a train, a gallery and a query.
Each subset has images for 750 identities with the resolution $128\times128$. 
The gallery and the query subsets share the same identities while having disjoint identities with the training set.
The training and the gallery sets have only 3 examples for each individual to simulate a limited data scenario. 
More examples per individual would make the re-identification task easier.
The query set has 5 images per individual. 

\subsubsection{Dataset of real images}

\begin{figure}
\begin{center}
 \includegraphics[width=\linewidth]{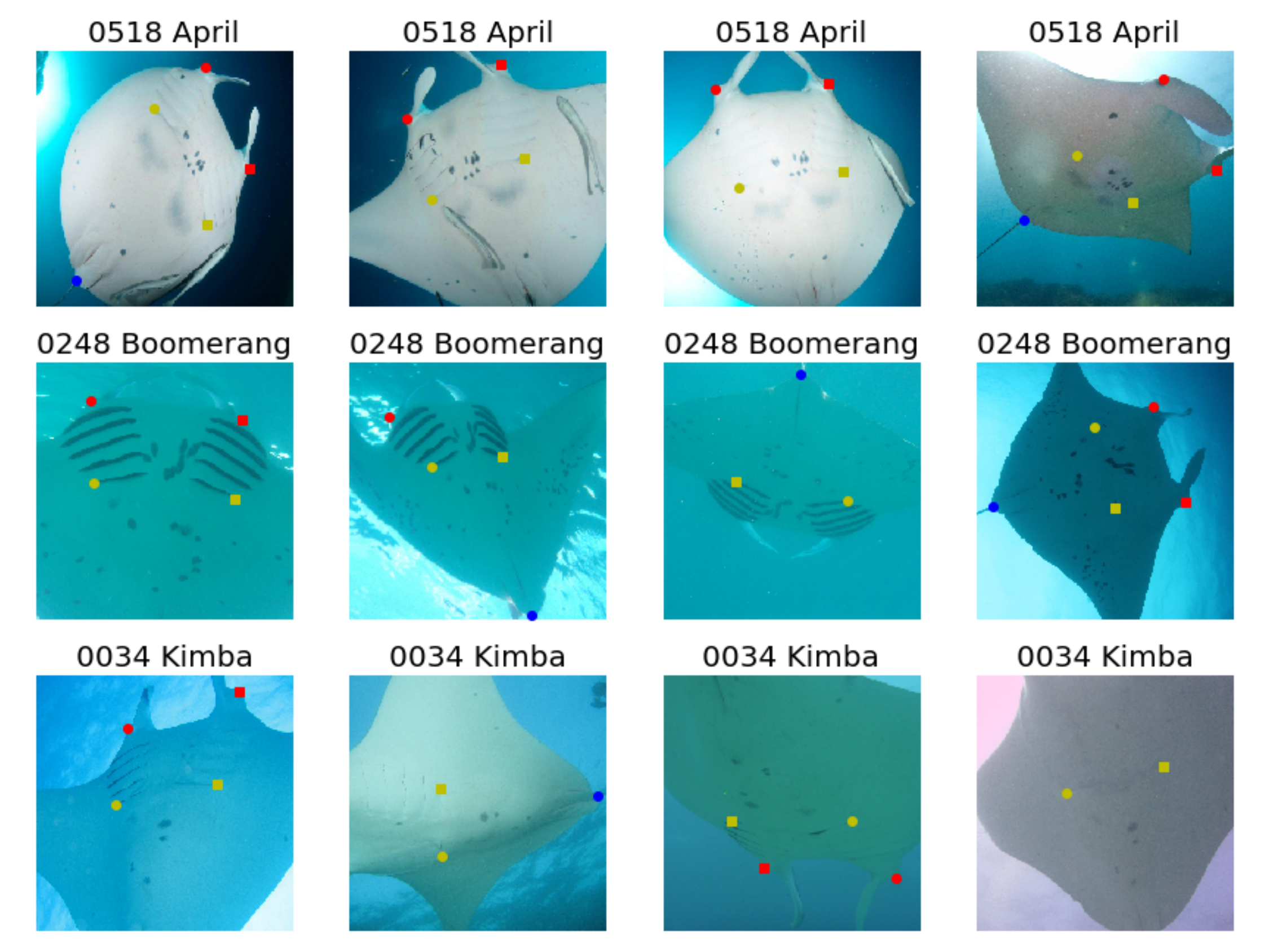}
\end{center}
   \caption{Example of three identities from a manta rays dataset (real images). Each row shows four examples of one identity. Coloured points are landmark locations and are plotted over images for illustration only and do not appear on images in the dataset.}
\label{fig:manta_identities}
\end{figure}

As a real dataset, we use images of manta rays collected by Project Manta (a multidisciplinary research program based at the University of Queensland, Brisbane, Australia). 
The dataset is challenging as images are captured underwater at oblique angles in different illumination conditions and with small occlusions (fish, water bubbles).
Each image has been manually annotated with five most distinctive landmarks: right eye, left eye, outer corner of the fifth right gill, outer corner of the fifth left gill, tail (see examples in Figure~\ref{fig:manta_identities}). 
We select eyes and a tail as landmarks as these are easy to identify in images.
Bottom gill slits on both sides have distal black marks that are salient and visible most times \cite{biology-manta}.
Only around half of the images have all 5 landmarks visible, 30\% of the images have 4 visible landmarks and the rest have 3 and less visible landmarks.

The training set has 110 identities with 1422 images in total.
The test set consists of 18 identities (different from the training) with 321 examples in total.
Images are taken by a large number of researchers and photographers so we assume that each image comes from a different camera.
Due to the limited size of the data we use one test set instead of a separate gallery and query sets.
The gallery set is created by combining the training set with two random images of each individual from the test set. The rest of the images from the test set are used as a query set. 
This way each query image has two matching examples in the gallery.

\subsection{Landmarks input as heatmaps}
Landmark coordinates are converted to heatmaps with one grayscale channel per a landmark, see Figure~\ref{fig:manta_heatmaps}.
The heatmap is created by running a Gaussian filter over a white disk on a black background to smooth the edges.
The center of the heatmap has equal intensity so there is no additional clue where the landmark is located. 
Heatmaps are used as an input to the model instead of exact coordinates to accommodate different levels of uncertainty in landmark locations.
If the landmark is not visible the heatmap is all zeros.

Heatmaps for the synthetic dataset are generated with three settings for the radius of the blob (5\%, 10\% and 20\% of the image size) to evaluate the sensitivity of the model to the uncertainty in landmark locations.
Heatmaps for the manta ray dataset have the radius 5\% of the image size.

\subsection{Model architecture}

We use ResNet50 model as a core feature extractor with the output feature maps pooled globally to produce a vector of size 2048.
Then one fully connected layer is used to reduce the dimension to 256.

The heatmap reconstruction block decodes heatmaps from an embedding using three blocks consisting of bi-linear upsampling with a factor of 2, a convolutional layer with  the kernel $3\times3$, a batch normalization layer and a relu activation function.
Reconstructed heatmaps have resolution $64\times64$ for any input size. 
This does not affect the network’s ability to reconstruct locations of body landmarks and allows us to minimise the number of parameters in the heatmap reconstruction branch.

\subsection{Training and evaluation}

Data augmentation is applied on the fly to images and corresponding heatmaps in the same way. 
We use rotations up to 360 degrees, zooming up to 20\% of image size and translations up to 20\%.
The same augmentation is applied when training the baseline model.
Heatmap augmentation NLA shifts the blob in heatmaps to imitate noise in landmark coordinates.
The minimal visible landmarks in MLA is set to 2 (out of possible 3) for the synthetic dataset and 3 (out of 5) for the manta ray dataset. The probability of missing a landmark is 50\%.

The model is trained on the training subset.
The test accuracy is obtained on new identities never seen during training.
The test accuracy is computed by retrieving predictions from the gallery set for each image in the query set.
We use top-1, top-5 and top-10 test accuracy for model evaluation.

\subsection{Results}
\subsubsection{Landmark-Id vs baseline}

The baseline results are obtained with only RGB images as input. 
\textit{Landmark-Id} model outperforms the baseline model on both synthetic and real datasets that demonstrates that additional pose information is beneficial for learning discriminative embeddings, see Tables~\ref{tab:poseid_synthetic_accuracy} and \ref{tab:poseid_manta_accuracy}.

\begin{figure}
\begin{center}
 \includegraphics[width=\linewidth]{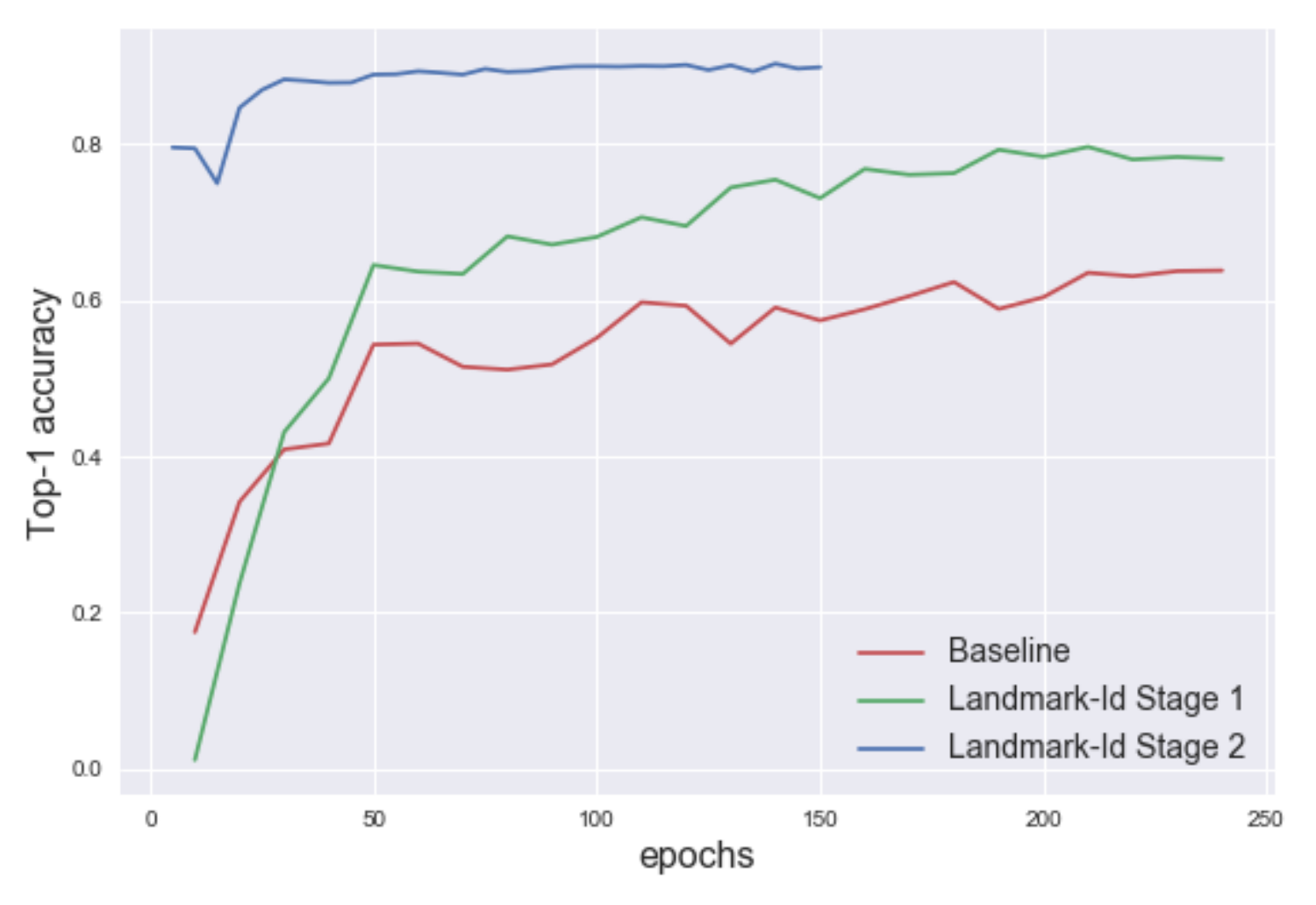}
\end{center}
   \caption{Progress of top-1 accuracy on the test set during training evaluated each 10 epochs on the synthetic dataset. \textit{Landmark-Id} Stage 2 continues training from Stage 1. Models are trained until convergence of the loss.}
\label{fig:training_plot}
\end{figure}

\begin{table}
\caption{\textit{Landmark-Id} outperforms the baseline re-identification model on the synthetic dataset. Stage 1 is the model trained with additional heatmap input and Stage 2 is the model with the heatmap reconstruction block.}
\label{tab:poseid_synthetic_accuracy}
\begin{center}
\begin{tabular}{|l|c|c|c|}
\hline
Model & Top-1 & Top-5 & Top-10 \\
\hline\hline
Baseline Reid & 63.81\% & 85.35\% & 90.94\% \\
Landmark-Id Stage 1 & 78.10\% & 91.82\% & 94.41\% \\
Landmark-Id Stage 2 & \textbf{89.53\%} & \textbf{95.96\%} & \textbf{96.98\%} \\
\hline
\end{tabular}
\end{center}
\end{table}

\begin{table}
\caption{Accuracy of re-identification on manta ray dataset. \textit{Landmark-Id} outperforms the baseline re-identification model. Stage 1 is the model trained with additional heatmap input and Stage 2 is the model with the heatmap reconstruction block.}
\label{tab:poseid_manta_accuracy}
\begin{center}
\begin{tabular}{|l|c|c|c|}
\hline
Model & Top-1 & Top-5 & Top-10 \\
\hline\hline
Baseline Reid & 44.00\% & 78.60\% & 84.70\% \\
Landmark-Id Stage 1 & 52.67\% & 80.11\% & 86.18\% \\
Landmark-Id Stage 2 & \textbf{62.04\%} & \textbf{89.82\%} & \textbf{91.96\%} \\
\hline
\end{tabular}
\end{center}
\end{table}

\textit{Landmark-Id} model reaches 89.53\% top-1 accuracy versus 63.81\% top-1 accuracy of baseline model on the synthetic data (Table~\ref{tab:poseid_synthetic_accuracy}).
The real data is more challenging.
Baseline model demonstrates 44.00\% top-1 accuracy while \textit{Landmark-Id} model goes up to 62.04\% (Table~\ref{tab:poseid_manta_accuracy}).

We evaluate top-1 accuracy on the test set during training every 10 epochs on the synthetic dataset, see Figure~\ref{fig:training_plot}.
\textit{Landmark-Id} model at Stage~1 (heatmap input with no reconstruction) shows higher accuracy than the baseline model. Stage~2 with auxiliary heatmap reconstruction further boosts the performance.
\textit{Landmark-Id} model without reconstruction block outperforms the baseline model.
Adding the heatmap reconstruction block is useful as it promotes usage of pose information during feature extraction and improves accuracy on both synthetic and real data.

The above results are obtained with no noise in landmark coordinates and heatmaps of 5\% of the image size. We analyse the sensitivity of the model to uncertainty and noise in the next section.

\subsubsection{Sensitivity to uncertainty in landmark locations}

\begin{table}
\caption{Sensitivity of \textit{Landmark-Id} to uncertainty in landmark locations is analysed with three sizes of heatmaps: 5\%, 10\% and 20\% of the image size. The model shows almost equal performance for heatmaps with the blob radius up to $\pm10\%$ of the image size.}
\label{tab:synth-sensitivity}
\begin{center}
\begin{tabular}{|l|c|c|c|}
\hline
Model & Top-1 & Top-5 & Top-10 \\
\hline\hline
Landmark-Id, hm 5\% & 86.13\% & 93.12\% & 95.82\% \\
Landmark-Id, hm 10\% & 84.72\% & 92.84\% & 95.31\% \\
Landmark-Id, hm 20\% & 66.62\% & 83.18\% & 88.85\% \\
\hline
\end{tabular}
\end{center}
\end{table}

We investigate the sensitivity of the model to uncertainty in landmark locations by training and evaluating the model on the synthetic dataset with different settings for the size of the bright blob in heatmaps.
Three experiments are conducted with the radius of the blob 5\%, 10\% and 20\% of the image size (see Figure~\ref{fig:heatmaps_size_noise}). NLA adds noise to heatmaps shifting the center from the actual landmark location.
The blob with a radius of $r\%$ means that the average noise in a landmark location is $\pm r\%$ of the image size.

Noise of 5\% and 10\% in landmark locations slightly decreases the accuracy (Table~\ref{tab:synth-sensitivity}).
The noise of $\pm20\%$ decreases the top-1 accuracy to 66.62\%.
This is a high level of uncertainty because the blob with the radius 20\% of the image size covers almost a quarter of the image.
We conclude that a landmark detection model should have at most 10\% error to predict landmark coordinates useful for re-identification.

\subsubsection{Sensitivity to missing landmarks}
\begin{table}
\caption{MLA (missing landmark augmentation) improves robustness of \textit{Landmark-Id} to not visible landmarks. Evaluated on real dataset of manta ray images.}
\label{tab:real-sensitivity-missing}
\begin{center}
\begin{tabular}{|l|c|c|c|}
\hline
Model & Top-1 & Top-5 & Top-10 \\
\hline\hline
Landmark-Id, with MLA &\textbf{62.04\%} & \textbf{89.82\%} & \textbf{91.96\%} \\
Landmark-Id, no MLA & 55.30\% & 87.47\% & 89.12\% \\
\hline
\end{tabular}
\end{center}
\end{table}

To evaluate the sensitivity of \textit{Landmark-Id} model to missing landmarks, we train the model without the MLA augmentation.
The synthetic dataset has most of the landmarks visible at all times so we use real data in this experiment.
The manta ray dataset has around 50\% images with all five landmarks visible, 30\% of images with four landmarks visible and the rest with three and less visible landmarks.

Without MLA augmentation top-1 accuracy drops to 55.30\% from 62.04\% on manta ray dataset (Table~\ref{tab:real-sensitivity-missing}).


\section{Conclusion}
We demonstrated that the additional input of body landmarks improves learning of discriminative embeddings.
This method is robust to uncertainty in landmark locations and tolerates errors in landmark coordinates up to 10\% of the image size.

We will conduct experiments on other real datasets (e.g., ATRW~\cite{tiger-dataset}, ELPephants~\cite{elephants-dataset}).
In the future, we plan to investigate how to train an accurate body landmark predictor on a small dataset and integrate it with the re-identification model.

{\small
\bibliographystyle{ieee}
\bibliography{egbib}
}

\end{document}